\documentclass[conference]{IEEEtran}
\IEEEoverridecommandlockouts
\usepackage{cite}
\usepackage{amsmath,amssymb,amsfonts}
\usepackage{graphicx}
\usepackage{textcomp}
\usepackage{xcolor}
\usepackage{booktabs}
\usepackage{mathtools}
\usepackage{bm}
\usepackage{amsmath}
\usepackage{algorithm}
\usepackage{algpseudocode}
\usepackage{subcaption}
\usepackage{multirow}
\usepackage[outline]{contour}
\usepackage{varwidth}
\usepackage{balance}

\def\BibTeX{{\rm B\kern-.05em{\sc i\kern-.025em b}\kern-.08em
    T\kern-.1667em\lower.7ex\hbox{E}\kern-.125emX}}
\begin{document}

\title{Federated Learning with Workload Reduction through Partial Training of Client Models and Entropy-Based Data Selection
}

\author{\IEEEauthorblockN{Hongrui Shi}
\IEEEauthorblockA{University of Sheffield \\
hshi21@sheffield.ac.uk}
\and
\IEEEauthorblockN{Valentin Radu}
\IEEEauthorblockA{University of Sheffield \\
valentin.radu@sheffield.ac.uk}
\and
\IEEEauthorblockN{Po Yang}
\IEEEauthorblockA{University of Sheffield \\
po.yang@sheffield.ac.uk}
}

\maketitle

\begin{abstract}
With the rapid expansion of edge devices, such as IoT devices, where crucial data needed for machine learning applications is generated, it becomes essential to promote their participation in privacy-preserving Federated Learning (FL) systems. The best way to achieve this desiderate is by reducing their training workload to match their constrained computational resources. 
While prior FL research has address the workload constrains by introducing lightweight models on the edge, limited attention has been given to optimizing on-device training efficiency through reducing the amount of data need during training. 
In this work, we propose \textbf{FedFT-EDS}, a novel approach that combines \textbf{F}ine-\textbf{T}uning of partial client models with \textbf{E}ntropy-based \textbf{D}ata \textbf{S}election to reduce training workloads on edge devices. By actively selecting the most informative local instances for learning, FedFT-EDS reduces training data significantly in FL and demonstrates that not all user data is equally beneficial for FL on all rounds. Our experiments on CIFAR-10 and CIFAR-100 show that FedFT-EDS uses only 50\% user data while improving the global model performance compared to baseline methods, FedAvg and FedProx. Importantly, FedFT-EDS improves client learning efficiency by up to 3 times, using one third of training time on clients to achieve an equivalent performance to the baselines. This work highlights the importance of data selection in FL and presents a promising pathway to scalable and efficient Federate Learning. 
\end{abstract}


\section{Introduction}

Federated Learning (FL)~\cite{mcmahan2017communication} has emerged as a prominent distributed learning paradigm that enables many distributed devices to collaborate in training machine learning models without sharing sensitive user data. At the same time, the widespread adoption of small edge computing devices, such as those in the Internet of Things (IoT), has led to an unprecedented surge in data generation -- much of it vital for advancing machine learning (ML) applications. However, training traditional models on resource-constrained devices, and the growing trend of adopting large foundation models across distributed systems, presents significant challenges to the efficiency and scalability of Federated Learning (FL). To address these challenges, there is a pressing need to optimize FL training processes to minimize the computational burden on edge devices. By reducing the workload of these devices, we can enable a broader range of resource-constrained devices to participate in FL, unlocking the potential of their untapped data sources.

While previous works~\cite{diao2020heterofl, horvath2021fjord, alam2022fedrolex, li2019fedmd, he2020group, nguyen2023enhancing} primarily focus on reducing the workload on devices by allowing clients to locally train a small part of the model or a subset of model parameters (submodel), the potential for enhancing on-device training efficiency by reducing the amount of training data required has been largely overlooked. 
In the space of centralized machine learning, active learning has been employed to prioritize data samples that can most effectively improve the model through additional training~\cite{li2024survey}. 
However, in the context of federated learning, active learning has been largely disregarded, primarily due to concerns about the perceived overhead it may impose on client workloads~\cite{li2021sample, nagalapatti2022your}.

In this work, we concurrently adopt two workload reduction strategies -- partial model training and data selection -- demonstrating their effectiveness in enhancing model training, accelerating convergence, and significantly reducing client workloads. These improvements enable federated learning across a wide range of devices.

Our proposed method, \textbf{Fed}erated \textbf{F}ine-\textbf{T}uning with \textbf{E}ntropy-based \textbf{D}ata \textbf{S}election (FedFT-EDS), leverages transfer learning to enable efficient federated fine-tuning. We begin by pre-training a global model on a large, accessible source domain to establish a strong foundation for feature extraction. During the federated learning process, clients are tasked with fine-tuning only a smaller portion of the model on their local data, significantly reducing the computational burden. To further optimize the training process, we introduce entropy-based data selection. Clients perform a single forward pass on their local data to identify a small subset of the most informative samples, which are then used to update the model. This strategy minimizes the computational overhead associated with data selection. To enhance the quality of the selected subset, we employ a hardened softmax function to prioritize samples with high uncertainty, ensuring that the model benefits from the most valuable training examples.

Our experiments on CIFAR-10 and CIFAR-100, conducted under non-IID data distribution, demonstrate that FedFT-EDS consistently outperforms popular FL baselines, FedAvg~\cite{mcmahan2017communication} and FedProx~\cite{li2020federatedoptimization}, and their variants with random data selection. FedFT-EDS achieves superior global generalization, improving performance by up to 5\%. Moreover, it significantly enhances learning efficiency, requiring less than one-third of the total training time compared to baselines. This reduction in client workload underscores the effectiveness of FedFT-EDS in optimizing FL. Notably, our results indicate that selecting 50\% of the most informative data, identified through entropy-based selection, yields better performance than using the entire dataset, highlighting the potential of strategic data selection in FL more broadly.

The main contributions of our research are:
\begin{itemize}
    \item Effective Data Selection: We introduce a novel entropy-based data selection method with hardened softmax activation to efficiently reduce the computational burden on clients in FL, particularly in non-IID settings.
    \item Fine-Tuning Strategy: We adopt a fine-tuning strategy that leverages a pre-trained global model to mitigate system and data heterogeneity, achieving improve performance compared to FL training from scratch.
    \item Data Heterogeneity Insights: Our experiments reveal that not all client data is equally beneficial for FL. Strategic data selection can significantly improve performance, even when using a subset of the data.
\end{itemize}

\section{Related Works}
\subsection{Partial Model Training}
Partial model training reduces workloads on clients by allowing them to train only a portion of the model rather than the full model. A series of FL works have proposed updating submodels of the global model on clients. These submodels are chosen to meet the local computational resources, thus supporting the participation of less capable devices. 

FjORD~\cite{horvath2021fjord} uses a dropout strategy to extract submodels from the global model, which are then broadcast for client update. The dropout probabilities, which are used to indicate how many units are disconnected from the global model, are chosen according to the computational capabilities of each client. The pruned units do not participate in the local updates, thus reducing the training cost on clients. In a similar study, HeteroFL~\cite{diao2020heterofl} reduces the size of each hidden layer of the global model through a determined ratio to extract its submodels. These ratios are determined based on the computational capability of each participating client. The server then aggregates the updated submodels keeping track of the position of each updated parameter in the global model. This concept works even for larger models, InclusiveFL~\cite{liu2022no} showing its efficacy on Transformer models. 

FedRolex~\cite{alam2022fedrolex} critiques FjORD and HeteroFL for unevenly training the parameters of submodels, which degrades global model performance under data heterogeneity. Instead, FedRolex uses a rolling window to extract a submodels between communication rounds, thus allowing each parts of the global model to be exposed to different local data distributions.  

Recent works~\cite{chen2019closer, tian2020rethinking} have found that fine-tuning a classifier on top of a fixed feature extractor, pretrained on a set of base tasks, can achieve performance comparable to state-of-the-art few-shot learning approaches. We apply this insight from centralized learning to the context of FL.

\subsection{Active Learning}
Active Learning (AL)~\cite{settles2009active, ren2021survey, tharwat2023survey} is a technique designed to reduce the labeling effort required for creating large training datasets by identifying high-value data points for labeling to improve training quality. 
It is based on the premise that machine learning algorithms can maintain their learning effectiveness using a carefully selected subset of training samples~\cite{ren2021survey}.
AL solutions based on uncertainty assessment~\cite{beluch2018power, liu2021influence, yuan2021multiple} are among the most popular techniques, characterized by their simplicity and low computational cost. 

Model prediction is often used as a proxy for data certainty, to determine the most uncertain data points for annotation and training.
Marginal sampling~\cite{scheffer2001active} and information entropy~\cite{settles2009active} are two widely used techniques for measuring data certainty. Notably, the latter selects samples with the highest entropy for annotation and training~\cite{joshi2009multi, luo2013latent, aggarwal2014active, li2019entropy}. Conversely, recent studies~\cite{li2022selecting, mindermann2022prioritized, yang2023towards} show that AL can enhance machine learning efficiency by reducing the size of the training dataset.

While AL has been extensively studied in centralized learning settings, its utility extends to federated learning, though it has received comparatively less attention. The few studies on FL that incorporate AL, such as LoGo~\cite{kim2023re} and F-AL~\cite{ahn2024federated}, focus on reducing the labeling burden on clients~\cite{jeong2020federated} rather than strategically selecting training data from the available and labeled local data. 

Li et al.~\cite{li2021sample} introduce the gradient upper bound norm on the global loss to calculate the importance scores of training samples. FLRD~\cite{nagalapatti2022your} learns a relevant data selector on the client side using reinforcement learning. The learned selector can identify local samples that are most beneficial for improving global model performance. 

\begin{figure*}[th]
  \begin{center}
    \includegraphics[width=0.9\linewidth]{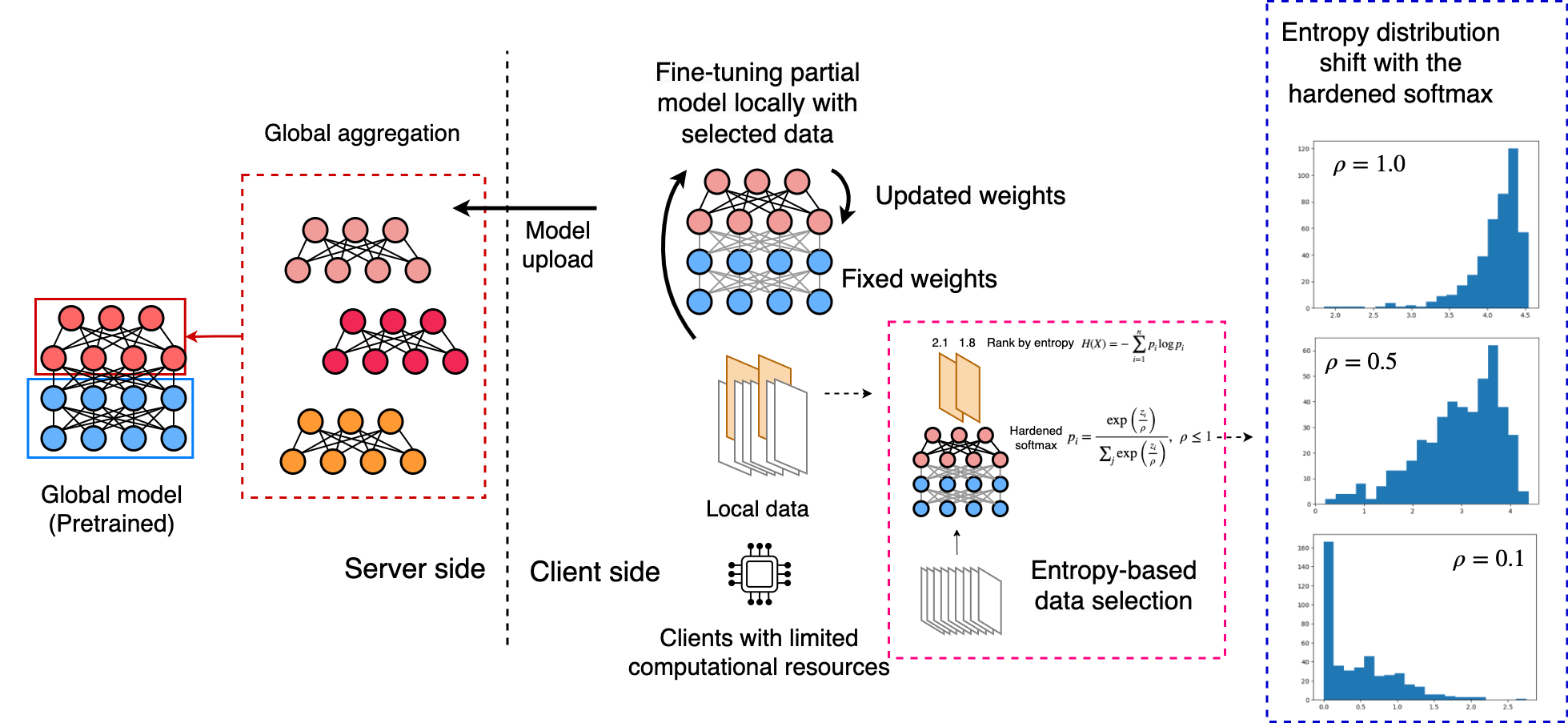}
  \end{center}
  \caption{The workflow of our proposed FedFT-EDS. A global pretrained model is split into two parts, one frozen feature extractor and one trainable upper part of the model. The entire model is shared with each client, but only the trainable upper part is updated. Clients also select their most valuable training samples by ranking them based on our hardened softmax activation.}
  \label{fig: method, entropy-based data selection, overview and entropy distribution}
\end{figure*}

FedEntropy~\cite{qian2023fedentropy} leverages entropy to select the most useful client models for global model aggregation. Each participating client computes the average entropy of its local data and uploads this value to the server. The server then eliminates clients that reduce the overall entropy, selecting a subset of clients that maximize global entropy. FedEntropy shows that combining models uploaded by this client model selection criteria improves global model performance. FedAvg-BE~\cite{orlandi2023entropy} also utilizes entropy, but for selecting batched local data to update client models. The batched local data comprises training samples with the highest entropy, which are most beneficial for learning. FedAvg-BE is the prior work most closely related to ours in its use of entropy for client data selection in FL. However, their focus is primarily on addressing the data heterogeneity challenge. In contrast, our work investigates the potential of entropy-based data selection for mitigating the straggler issue. Further, we argue that batch level entropy masks the utility of individual samples, so we focus on selecting the best individual samples for training by separately calculating entropy at sample level. Another distinguishing factor is our introduction of the hardened softmax activation to enhance the relevance of data selection by entropy.

Our novel approach combines two client workload reduction strategies, data selection and partial model training, tackling this important challenge in heterogeneous FL from both directions simultaneously.

\section{Proposed Method}\label{sec: method}
Figure~\ref{fig: method, entropy-based data selection, overview and entropy distribution} presents the workflow and Algorithm~\ref{alg: FedFT} the implementation of our proposed \textbf{Fed}erated \textbf{F}ine-\textbf{T}uning with \textbf{E}ntropy-based \textbf{D}ata \textbf{S}election (FedFT-EDS).

\begin{algorithm}
\caption{\textbf{FedFT-EDS}: \textbf{Fed}erated \textbf{F}ine-\textbf{T}uning with \textbf{E}ntropy-based \textbf{D}ata \textbf{S}election.}

\begin{algorithmic}[1]
\State \begin{varwidth}[t]{0.95\linewidth}\textbf{Pretraining Phase:} Pretrain the global model on the source domain, initialise the global model to $w^{1}_g = \{\phi, \: \theta^{1}_{g}\}$, every client and the server keep a copy of $\phi$. \end{varwidth}
\vspace{2mm}
\State \textbf{Input:} total $T$ rounds, $E$ local updates epochs, initialised $w^{1}_g$, total $N$ clients. 
\For{$t=1, \ldots, T$}
    \State \begin{varwidth}[t]{0.9\linewidth}{$K$ random clients are available for training and they download the upper part of the global model $\theta^{t}$ from the server.} \end{varwidth}
    \vspace{2mm}
    
    \State \textbf{Clients:}
    \For{Client $k \in [1, K]$}
        \State \begin{varwidth}[t]{0.8\linewidth} {$\mathcal{D}^{t}_{k, select} \leftarrow$ \textbf{Data Selection} $\left(; \theta^{t}, \phi, \mathcal{D}_{k} \right)$ with Equation~\ref{eq: FedFT, local updates for calculating predictions, chapter 6} and Equation~\ref{eq: FedFT, local updates for calculating entropoy, chapter 6}.} \end{varwidth}
        \vspace{2mm}
        \State \begin{varwidth}[t]{0.8\linewidth}{$\theta^{t+1} \leftarrow$ \textbf{Local updates} $\left(\theta^{t}; \: \phi, \mathcal{D}^{t}_{k, select}, E\right)$ with Equation~\ref{eq: FedFT, locally update the upper part with selected data}.}\end{varwidth}
        \vspace{2mm}

    \State Client $k$ uploads $\theta^{t+1}_{k}$ to the server.
    \EndFor
    \vspace{2mm}
    \State \textbf{Server:}
    \State \begin{varwidth}[t]{0.9\linewidth}{Collect local updates and compute the upper part of the global model to $\theta^{t+1}_{g}$ with Equation~\ref{eq: FedFT global updates}.}\end{varwidth}
    \State \begin{varwidth}[t]{0.9\linewidth}{Form the global model $w^{t+1}_{g} = \{\phi, \; \theta^{t+1}_{g}\}$ to start the next iteration.} \end{varwidth}
    \vspace{2mm}
    \State \textbf{Return} Global model $w^{t}_{g} = \{\phi,\; \theta^{t}_{g}\}$
\EndFor
\State \textbf{Return} Global model $w^{T}_{g} = \{\phi, \;\theta^{T}_{g}\}$
\end{algorithmic}

\label{alg: FedFT}
\end{algorithm}

\subsection{Preliminaries}\label{sec: method, FedFT-EDS}

We introduce the standard FL problem setup for the preliminaries. For a client indexed by $k$, its local training dataset is denoted as 
$\mathcal{D}_{k}$, comprising a set of samples represented as $\{(x_k^{(i)}, \: y_k^{(i)})\}^{\left|\mathcal{D}_{k}\right|}_{1}$, where $x_k^{(i)}$ and $y_k^{(i)}$ are the $i$-th local instance and its corresponding label respectively. The model trained in FL is called the global model, denoted as $M_g$, whose parameters are denoted as $w_g$. The learning objective is to find the $w_g$ that minimises the combined local losses across all clients, as described by Equation~\ref{eq: FL objective} below. 

\begin{equation}
    \underset{w_g}{\arg \min } \mathcal{L}(w_g)=\sum_{k=1}^N p_{k} L_k(w_g)
\label{eq: FL objective}
\end{equation}

\noindent where, $N$ is the size of the client pool, $L_k(w_g)$ is the empirical local loss. $p_{k}=\frac{\left|\mathcal{D}_{k}\right|}{|\mathcal{D}|}$ is the coefficient to weight individual losses, determined by the proportion of local data relative to the total client data $\mathcal{D} \triangleq \bigcup_{k \in[N]} \mathcal{D}_k$. 

\subsection{Pretraining the Global Model}
FL benefits significantly from pretraining a global model in terms of generalisation, convergence, and fairness~\cite{nguyen2022begin}. Our proposed entropy-based data selection method adopts the pretraining strategy for the initial global model prior to starting the FL update rounds. Concretely, the global model is first pretrained on a source domain that is assumed available on the server. Then the model is distributed to all clients for federated learning on a downstream task. 

We take advantage of this global model pretraining strategy to reduce the training workload on clients. It is common practice in centralised learning to pretrain a model on large datasets from a source domain, and once the feature extractor part of the model has gained enough skill, to freeze parts of the model and continue to fine-tune it on data from downstream task. We adopt this approach in our FL approach. Here, clients are tasked with fine-tuning a small part of the pretrained model, rather than the entire model as in general FL, significantly reducing the training cost. We provide further insights for our choice through empirical observations in the experiments section. 

\subsection{Local Updates of FedFT-EDS} \label{sec: method, local updates of FedFT-EDS}
FedFT-EDS performs local updates by fine-tuning a well-defined part of the model with local instances actively selected based on their entropy.

Formally, at communication round $t$, client $k$ first downloads the global model as $M^{t}_k$ parameterised with $w^{t}_k$. The client then selects local instances for training $M^{t}_k$. To achieve this, the client model performs a forward pass with all available client data $x_k^{(i)} \in \mathcal{D}_{k}, \;i = 1, \cdots, {|\mathcal{D}_{k}|}$ to obtain their softmax activations as follows:

\begin{equation}
P_k^{t, (i)} = f_{w^{t}_k}(w^{t}_k ; \; x_{k}^{(i)})
\label{eq: FedFT, local updates for calculating predictions, chapter 6}
\end{equation}

\noindent where $P_k^{t, (i)}$ is the probability vector output by the softmax activation layer. Then the Shannon entropy for an instance $x_k^{(i)}$ is calculated with Equation~\ref{eq: FedFT, local updates for calculating entropoy, chapter 6}.

\begin{equation}
H_k^{t, (i)} = -\sum_{p^{t, (i)}_j \in P_k^{t, (i)}} p^{t, (i)}_j \log p^{t, (i)}_j
\label{eq: FedFT, local updates for calculating entropoy, chapter 6}
\end{equation}

\noindent where $p^{t, (i)}_j$ is the probability of labeling the instance $x_k^{(i)}$ as the $j$-th class in the possible outcomes $\{c_1, c_2, \dots, c_n\}$. 

The client $k$ ranks instance $x_k^{(i)} \in \mathcal{D}_{k}$, \mbox{$i = 1, \cdots, {|\mathcal{D}_{k}|}$} based on the calculated entropy values. High entropy implies that the model is more uncertain about its prediction and vice versa. Therefore, samples with higher entropy are seen as harder but more valuable ones for learning than those with lower entropy. By ranking samples according to their entropy values, the client identifies a small subset of local data, $\mathcal{D}^{t}_{k, select}$, which contains instances with higher entropy values that can update $w^{t}_k$ efficiently. 

Using the pretrained global model, client $k$ only fine-tunes the upper part of $M^{t}_k$ on $\mathcal{D}^{t}_{k, select}$ while keeping the lower part frozen. Specifically, the parameters of the client model are denoted as $w^{t}_{k} = \{\phi, \: \theta^{t}_{k}\}$, where $\phi$ is the feature extractor at the bottom of the model and $\theta^{t}_{k}$ denotes the upper part. Here, $\phi$ is not indexed with the communication round $t$ nor with the client index $k$ because it comes from the pretrained global model and all clients have the same copy for that part of the model. 
Thus, only $\theta^{t}_{k}$ is updated to $\theta^{t+1}_{k}$ using data $\mathcal{D}^{t}_{k, select}$:

\begin{equation}
    \theta^{t+1}_k \leftarrow \theta^{t}_k -\lambda \nabla_{\theta^{t}_k} \ell_k(\theta^{t}_k ;\phi, \mathcal{D}^{t}_{k, select}),
    \label{eq: FedFT, locally update the upper part with selected data}
\end{equation}

\noindent where $\lambda$ is the learning rate. The client $k$ completes its local update and uploads the trained part of the model (upper part of the model, $\theta^{t+1}_{k}$) to the server.

\subsection{Global Updates of FedFT-EDS}

FedFT-EDS builds on FedAvg to update the global model. Specifically, FedFT-EDS fuses the updated model parameters $\theta^{t+1}_{k}, k \in \{1, \dots, K\}$ as follows:

\begin{equation}
    \theta^{t+1}_g \leftarrow \sum_{k=1}^K p^{t}_{k} \theta^{t+1}_k
\label{eq: FedFT global updates}
\end{equation}

\noindent where, $p^{t}_{k}$ is calculated based on the selected client data $p^{t}_{k}=\frac{\left|\mathcal{D}^{t}_{k, select}\right|}{|\mathcal{D}^{t}|}$ with $\mathcal{D}^{t} \triangleq \bigcup_{k \in[K]} \mathcal{D}^{t}_{k, select}$. The server constructs the global model for the next communication round $t+1$ from the frozen part and from the updated part $w^{t+1}_g = \{\phi, \: \theta^{t+1}_{g}\}$ and distributes it to the clients. 

As discussed in local updates, the feature extractor $\phi$ is not updated during FL iterations. Hence, the server and the clients only need to communicate the upper part of the global model regularly, $\theta^{t}_g$, which also reduces the communication overhead in FL. Algorithm~\ref{alg: FedFT} describes the details of the proposed FedFT-EDS.

\subsection{Entropy Calculation with Hardened Softmax}\label{sec: method hardened softmax}
In this section, we introduce the hardened softmax to enhance the efficacy of our data selection method. For a given local sample, the more confident the model prediction is (lower entropy), the less it contributes to learning. The hardened softmax increases the entropy of such confident samples, effectively excluding them from local updates, thus ensuring that only highly uncertain samples are selected for training.

Equation~\ref{eq: FedFT, local updates for calculating predictions, chapter 6} and Equation~\ref{eq: FedFT, local updates for calculating entropoy, chapter 6} describe how the entropy is calculated to support the local sample selection. However, a key limitation of Shannon entropy is that small changes in $p^{t, (i)}_j$ within the probability vector result in only minor changes in entropy~\cite{li2019entropy}. This characteristic can restrict the effectiveness of Shannon entropy in identifying the most uncertain instances. 
Specifically, a slight increase in $p^{t, (i)}_j$ indicates that the model has become slightly more confident in classifying the instance into category $j$. Since the model gains less from training on instances it is already confident about, it would be preferable for a small increase in $p^{t, (i)}_j$ to cause a significant decrease in entropy, thereby excluding the instance from local updates.

While we cannot alter the Shannon entropy defined by Equation~\ref{eq: FedFT, local updates for calculating predictions, chapter 6}, we can reshape the probability distribution in the probability vector in Equation~\ref{eq: FedFT, local updates for calculating predictions, chapter 6} by amplifying the change in $p^{t, (i)}_j$. 
To address this limitation of entropy-based data selection, we employ the hardened softmax activation. The hardened softmax is a variant of the softmax function, parameterized by a temperature $\rho$ set to a value greater than 1, as defined in Equation~\ref{eq: softmax with temp}.

\begin{equation}
p_{i}=\frac{\exp \left(\frac{z_{i}}{\rho}\right)}{\sum_{j} \exp \left(\frac{z_{i}}{\rho}\right)}
\label{eq: softmax with temp}
\end{equation}

Our hardened softmax is inspired by the softened softmax used in knowledge distillation (KD)~\cite{hinton2015distilling}. In KD, the temperature parameter $\rho$ introduced in the softmax activation is set to a value greater than 1, producing a softer probability distribution that enriches the information transferred between the teacher and student models by aligning their output spaces. We adapt this concept for entropy-based data selection by setting $\rho$ to a value smaller than 1, effectively `hardening' the probability distribution. This adjustment ensures that a slight increase in prediction confidence triggers a significant decrease in entropy, leading to that instance not being selected for the learning round.

We demonstrate that the entropy distribution shifts significantly when adjusting the temperature parameter ($\rho$) in the softmax function. Figure~\ref{fig: method, entropy-based data selection, overview and entropy distribution} illustrates this shift with three settings for $\rho$ (1.0, 0.5, and 0.1). The entropy distribution is generated by inputting the local instances of a client (from CIFAR-100) into a pretrained neural network. For lower values of $\rho$ ($\rho=0.1$), the distribution becomes concentrated in the lower entropy range, with only a narrow tail extending into the higher entropy region. This makes data points with the highest entropy more distinguishable. Conversely, when $\rho$ is set to a larger value ($\rho=1.0$), the higher entropy region becomes densely populated, making it harder to identify the most uncertain and useful instances effectively.

\section{Experiments}

\subsection{Experimental setup}
\subsubsection{Datasets and models}
 We evaluate FedFT-EDS on two image classification tasks using CIFAR-10, CIFAR-100~\cite{Krizhevsky2009} and a speech classification task using the Google Speech Command dataset~\cite{warden2018speech}. The global model is pretrained on the ImageNet Small $32 \times 32$ dataset prior to FL. The number of local update epochs $E$ is set to 5. SGD optimiser with a learning rate of 0.1 and momentum of 0.5 is used for the local updates. The temperature in the hardened softmax activation is set to 0.1. The Wide ResNet (WRN) model~\cite{zagoruyko2016wide} with the depth of 16 and width of 1 is used in our FL experiments. The client model is fine-tuned from layer 3, with layer 1 and layer 2 being fixed during local updates. Client data heterogeneity is also simulated in our experiments. Following many prior works~\cite{hsu2019measuring, lin2020ensemble, he2020fedml}, the Dirichlet distribution, denoted by $Diri(\alpha)$, is employed to partition the non-IID client data. A small $\alpha$ value suggests strong data heterogeneity and vice versa. In our experiments, we use $Diri(0.1)$ and $Diri(0.5)$ to simulate different levels of data heterogeneity. In Section~\ref{sec: ablation study}, we further conduct ablation studies on the effects of temperature values, model layer levels for fine-tuning, and data heterogeneity with different setup.

\subsubsection{Baselines}
 We choose two popular FL baselines, FedAvg and FedProx, to compare with our FedFT-EDS. FedAvg is the standard FL baseline that we introduced in the preliminary section of Section~\ref{sec: method}. FedProx advances FedAvg to tackle the model shift problem by using a proximal term that prevents the local updates from drifting far from the global objective. Additionally, we construct two new baselines by modifying them to use random data selection for local updates, denoted as FedAvg-RDS and FedProx-RDS, which are used to demonstrate the effect of reduced training data. Basically, FedAvg-RDS and FedProx-RDS clients randomly select a proportion of their data to update the global model locally during each round. Finally, we use the FedFT-RDS to denote the baseline that adopts the same partial model training strategy to FedFT-EDS but selected client data randomly, contrasting with the entropy-based data selection. 
\subsubsection{Setup of random data selection}
 Section~\ref{sec: method, local updates of FedFT-EDS} presented how FedFT-EDS performs the data selection from local data by relying on entropy information at the beginning of each local round. This is due to entropy calculated over the model output on each data instance changing with the model update. As such, FedFT-EDS performs the data selection dynamically. For fair comparison, we allow clients to perform a uniform random data selection from their local data at the start of each local update. Thus, the client data used for model updates varies between rounds. 

\subsection{Benefits of Pretraining} \label{sec: method, pretraining}

This section reports our empirical studies on the benefits of pretraining. We show that the pretrained global model is robust to the model shift problem induced by the data heterogeneity, leading to improved global model performance in FL. In this study, we use CIFAR-100 and Small ImageNet as the source domains separately to pretrain the global model and perform FL on CIFAR-10 with 10 clients holding heterogeneous data.

Table~\ref{tab: data selection, transfer learning, preliminary, 10 clients.} reports FL performance on the test set of CIFAR-10. Using FedAvg as the baseline, we compare the global model performance of using the pretraining strategy and that of without using pretraining. We clearly see that pretraining on a source domain, either from CIFAR-100 or from Small ImageNet, significantly improves the global model performance by up to 8\%. Pretraining with Small ImageNet yields better results due to pretraining exposing the model to broader diversity and richer samples.

\begin{table}
\caption{Pretraining improves FL performance (top-1 accuracy (\%)) on the downstream task.}
\centering
\begin{tabular}{@{}lllll@{}}
\toprule
\textbf{Method} & \textbf{Model} & \textbf{Pretraining} & \boldsymbol{$Diri(0.1)$} & \boldsymbol{$Diri(0.5)$}  \\ \midrule
\multirow{3}{*}{FedAvg} & \multirow{3}{*}{WRN} & na &  67.46 & 79.53 \\ 
 &  & CIFAR-100 &  70.46 \contour{black}{$\uparrow$} & 80.70 \contour{black}{$\uparrow$}\\ 
 &  & Small ImageNet &  75.18 \contour{black}{$\uparrow$} & 81.73 \contour{black}{$\uparrow$}\\ 
\bottomrule
\end{tabular}

\label{tab: data selection, transfer learning, preliminary, 10 clients.}
\end{table}

Further, we observe the performance gap under different levels of data heterogeneity, seeing that pretrained global model has clear advantage in strong data heterogeneity conditions. Pretraining improves FedAvg by around 8\% for the case of $Diri(0.1)$, compared to just 2\% improvement for $Diri(0.5)$. These insights indicate that pretraining makes client models robust to model shift introduced by client data heterogeneity. We use the Centred Kernel Alignment {(CKA)~\cite{kornblith2019similarity, nguyen2020wide}} to visualize this robustness of the pretrained model in FL.

CKA is a widely used metric to observe the similarity of neural networks by comparing their latent representations. A high CKA score indicates that the compared models learn close latent representations. CKA can serve as a metric for measuring the magnitude of model shifts. Specifically, we use CKA to measure the similarity of updated models across clients. Given heterogeneous client data, locally updated models tend to deviate from each other, resulting in degraded FL performance. This deviation cause differences in the latent space of the client-updated models, reflected on the low CKA score.  

With the same experimental setup described above, we calculate the CKA score for all combinations of paired client-updated models. CKA is computed at three different layer levels inside the WRN model, layers $l \in \{low, mid, up\}$, by inferring all the instances in CIFAR-10 test set. Figure~\ref{fig: CKA heatmaps visualise model similarity, Diri(0.1)} and Figure~\ref{fig: CKA heatmaps visualise model similarity, Diri(0.5)} present the heatmaps of CKA scores produced by the updated models of 10 clients with data heterogeneity of $Diri(0.1)$ and $Diri(0.5)$ respectively. An entry $(i, j)$ in the heatmap is the averaged CKA score over the inferences with CIFAR-10 test set between client $i$ and client $j$.

\begin{figure}
  \centering
    \begin{subfigure}{.45\linewidth}
    \centering
    \centerline{\includegraphics[width=\linewidth]{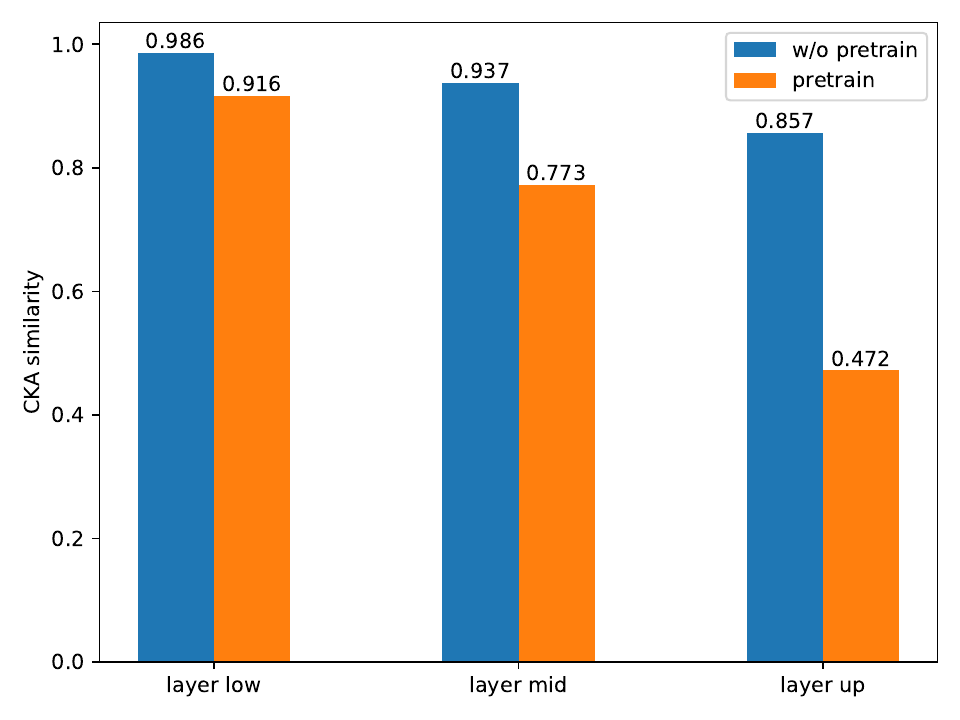}}
    \caption{$Diri(0.1)$}
    \end{subfigure} 
    \begin{subfigure}{.45\linewidth}
    \centering
    \centerline{\includegraphics[width=\linewidth]{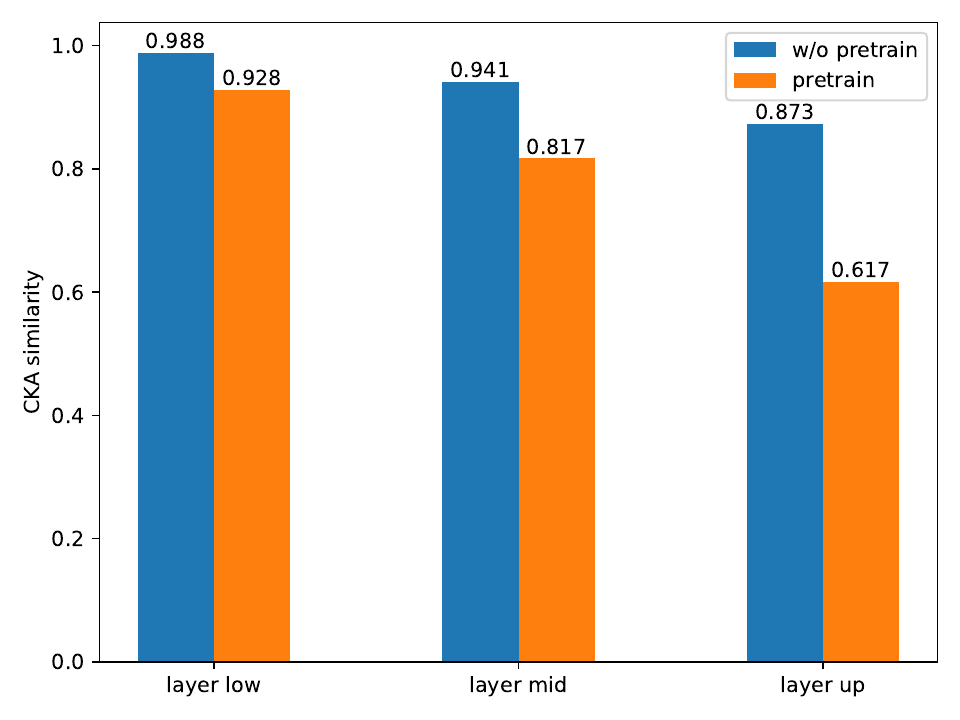}}
    \caption{$Diri(0.5)$}
    \end{subfigure} 
    
  \caption{Averaged CKA similarity at different layer levels across the locally trained models. }
  \label{fig: CKA bar chart for similarity comparison}
\end{figure}

Similarities between client-updated models are notably higher with darker entries shown in the charts when pretraining on Small ImageNet across all three layer levels (second row of charts in Figure~\ref{fig: CKA heatmaps visualise model similarity, Diri(0.1)} and Figure~\ref{fig: CKA heatmaps visualise model similarity, Diri(0.5)}), indicative of less model shift. Figure~\ref{fig: CKA bar chart for similarity comparison} shows a more compact representation of the same CKA similarity averaged over the scenarios $Diri(0.1)$ and $Diri(0.5)$. The gap between using pretraining and without pretraining is more severe for $Diri(0.1)$.

\begin{figure}
  \centering
    \begin{subfigure}{.32\linewidth}
    \centering
    \centerline{\includegraphics[width=\linewidth]{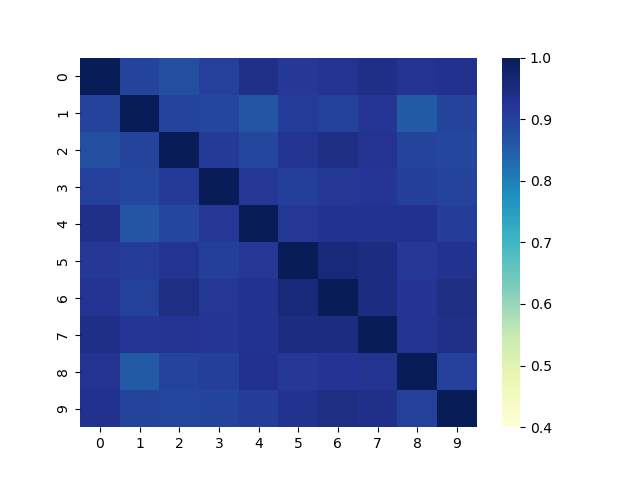}}
    \caption{low}
    \end{subfigure} 
    \begin{subfigure}{.32\linewidth}
    \centering
    \centerline{\includegraphics[width=\linewidth]{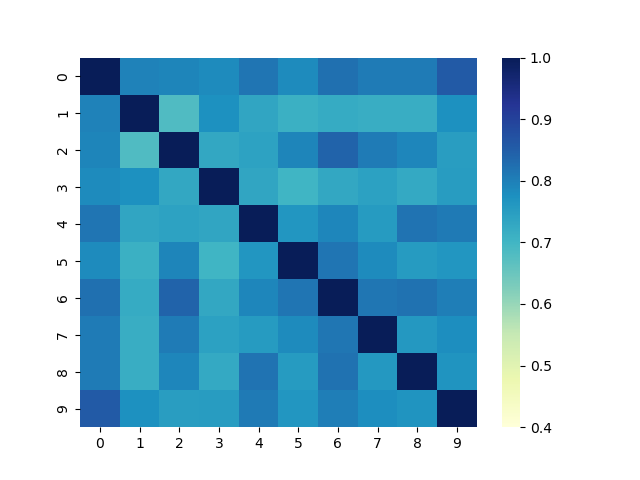}}
    \caption{mid}
    \end{subfigure} 
    \begin{subfigure}{.32\linewidth}
    \centering
    \centerline{\includegraphics[width=\linewidth]{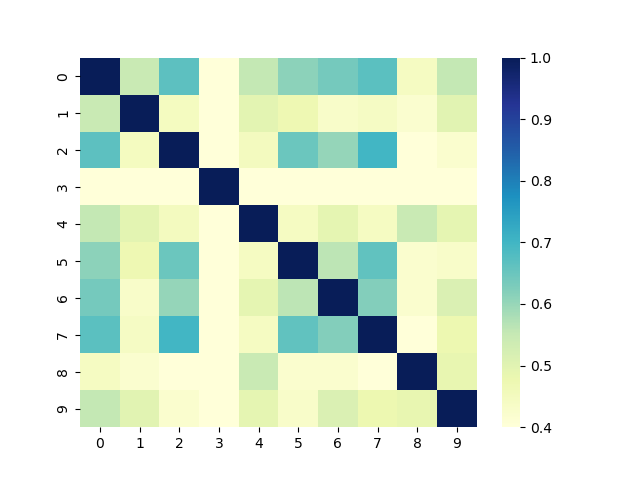}}
    \caption{up}
    \end{subfigure}
    
    \begin{subfigure}{.32\linewidth}
    \centering
    \centerline{\includegraphics[width=\linewidth]{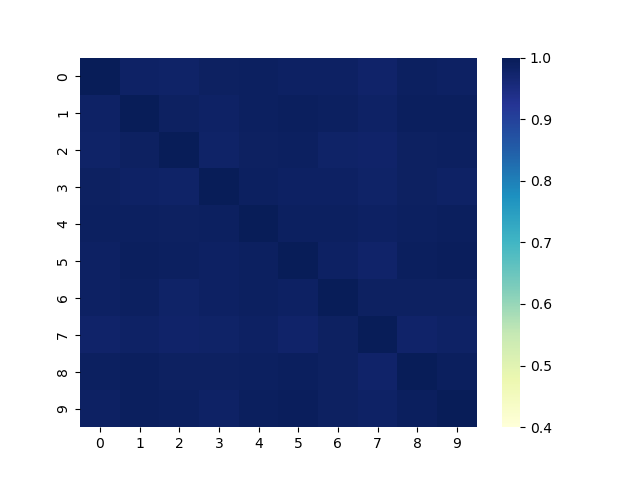}}
    \caption{pretrain, low}
    \end{subfigure} 
    \begin{subfigure}{.32\linewidth}
    \centering
    \centerline{\includegraphics[width=\linewidth]{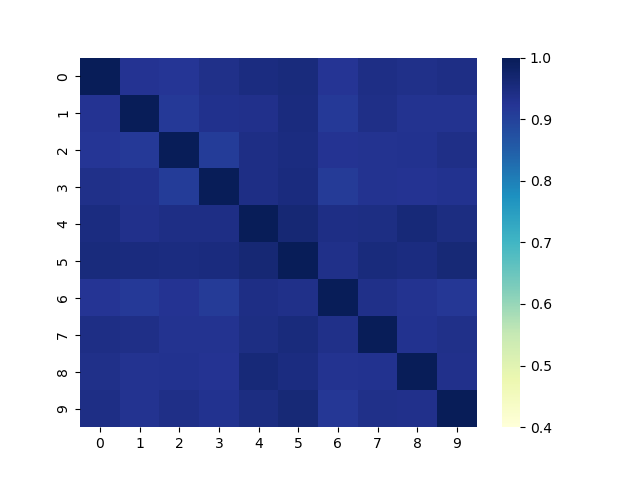}}
    \caption{pretrain, mid}
    \end{subfigure} 
    \begin{subfigure}{.32\linewidth}
    \centering
    \centerline{\includegraphics[width=\linewidth]{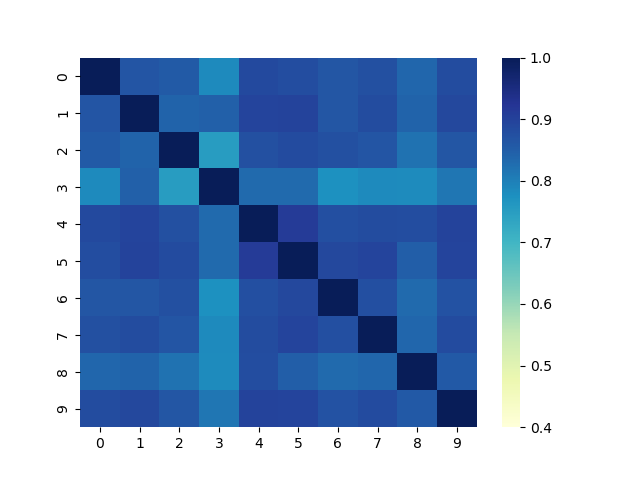}}
    \caption{pretrain, up}
    \end{subfigure}

  \caption{Heatmaps of the CKA similarity in the scenario of 10 clients and $Diri(0.1)$. A darker entry implies a higher similarity between the paired models indexed by the coordinate, suggesting they are less deviated from each other on heterogeneous data.}
  \label{fig: CKA heatmaps visualise model similarity, Diri(0.1)}
\end{figure}

\begin{figure}
  \centering
    \begin{subfigure}{.32\linewidth}
    \centering
    \centerline{\includegraphics[width=\linewidth]{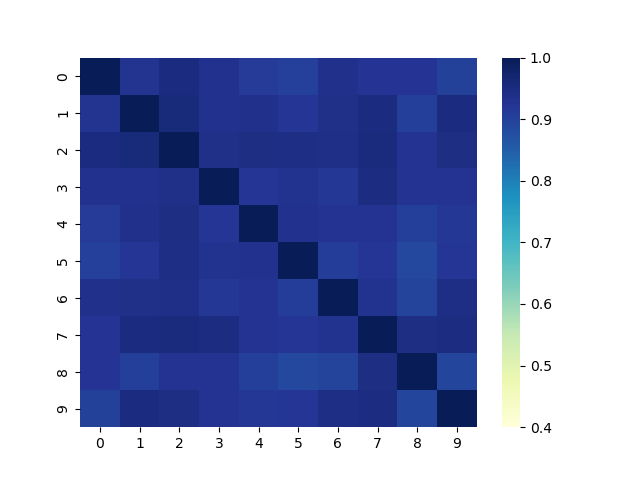}}
    \caption{low}
    \end{subfigure} 
    \begin{subfigure}{.32\linewidth}
    \centering
    \centerline{\includegraphics[width=\linewidth]{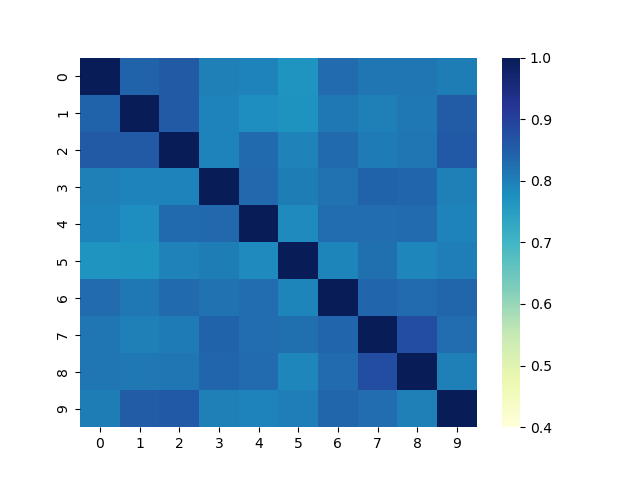}}
    \caption{mid}
    \end{subfigure} 
    \begin{subfigure}{.32\linewidth}
    \centering
    \centerline{\includegraphics[width=\linewidth]{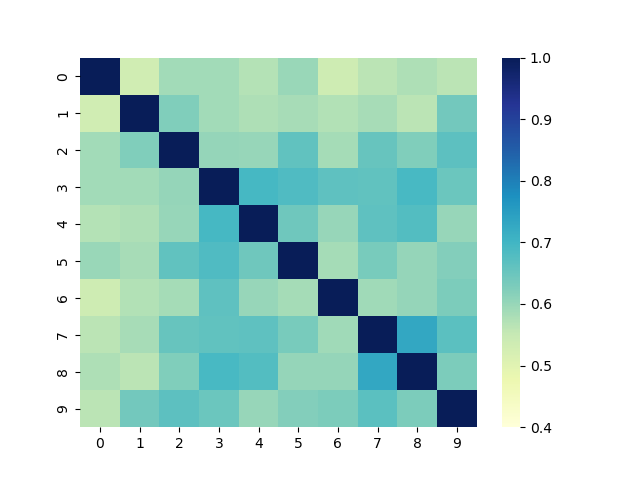}}
    \caption{up}
    \end{subfigure} 
    
    \begin{subfigure}{.32\linewidth}
    \centering
    \centerline{\includegraphics[width=\linewidth]{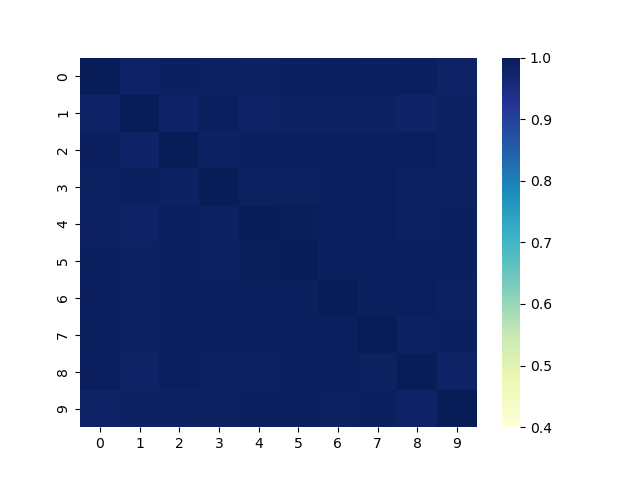}}
    \caption{pretrain, low}
    \end{subfigure} 
    \begin{subfigure}{.32\linewidth}
    \centering
    \centerline{\includegraphics[width=\linewidth]{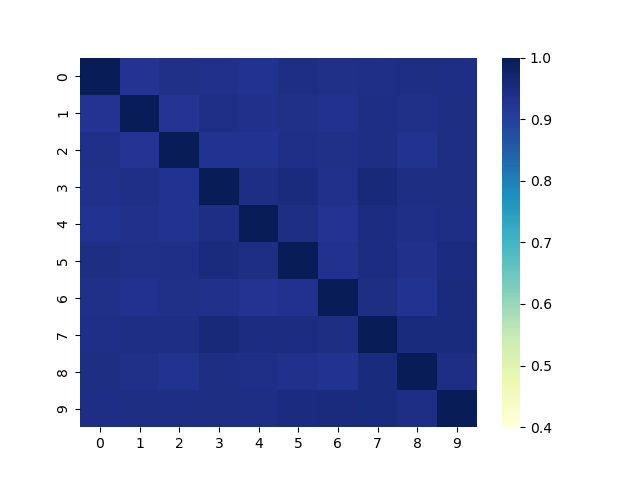}}
    \caption{pretrain, mid}
    \end{subfigure} 
    \begin{subfigure}{.32\linewidth}
    \centering
    \centerline{\includegraphics[width=\linewidth]{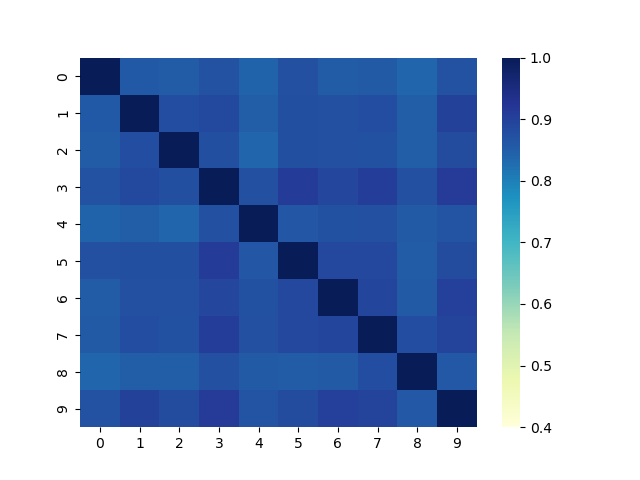}}
    \caption{pretrain, up}
    \end{subfigure} 
    
  \caption{Heat maps of the CKA similarity in the scenario of 10 clients and $Diri(0.5)$. }
  \label{fig: CKA heatmaps visualise model similarity, Diri(0.5)}
\end{figure}

\subsection{Close-Domain Evaluation}
This section introduce our experiment results when the FL task is a close domain to the pretraining domain. Notably, pretraining the global model on Small Imagenet, Table~\ref{tab: local data selection CIFAR-10/100, 10 clients.} reports the performance of FedFT-EDS and baseline methods on image classification, CIFAR-10 and CIFAR-100. The chosen baselines are FedAvg, FedProx, FedAvg-RDS, FedProx-RDS, and FedFT-RDS. In addition, we include FedAvg without global model pretraining (the vanilla FedAvg trained in FL from scratch), and the centralised training to anchor the results. FedAvg (scratch) and centralised training are the lower and the upper bound of performance respectively. Regarding the data selection, local instances are selected in proportion to the available local data, defined by $P_{ds}$. Where $P_{ds}$ is 100\%, no data selection is performed, using all local data for local updates. For FedFT-EDS, FedFT-RDS, FedAvg-RDS, we select fewer local samples by setting $P_{ds}$ to 10\%. 

Table~\ref{tab: local data selection CIFAR-10/100, 10 clients.} shows that FedFT-EDS outperforms all the other baselines expect for the centralised training. These results are insightful in three ways. First, pretraining significantly improves the global model performance. The most substantial boost is observed in strong data heterogeneity, for $Diri(0.1)$. Second, both FedFT-RDS and FedFT-EDS outperform baselines without using partial local fine-tuning by large margins, up to 5\% on CIFAR-10 and up to 3\% on CIFAR-100. The superior performance achieved by FedFT-RDS over FedAVG-RDS shows that fine-tuning the upper part of the client model is more effective than updating the entire model using all the available data. The combination of pretraining and partial local fine-tuning significantly closes the performance gap between FL and centralised training by 30\% to 75\%. Finally, the proposed entropy-based data selection has an edge over the random data selection. FedFT-EDS improves the performance of FedFT-RDS by 0.4\% to 2.7\%, demonstrating that sampling useful training instances with entropy information is superior. 

\begin{table}
\caption{Model performance measured on the global model for our FedFT-EDS and the baselines with full participation of 10 clients. All methods, except FedAvg w/o pretraining (pt), adopt the pretraining strategy. $\alpha$ value sets up the Dirichlet distribution for non-IID client data.}
\centering
\begin{tabular}{@{}llllll@{}}
\toprule
\multirow{2}{*}{\textbf{Method}}& \multirow{2}{*}{\boldsymbol{$P_{ds}$}}   & \multicolumn{2}{c}{\textbf{CIFAR-10}} & \multicolumn{2}{c}{\textbf{CIFAR-100}}  \\ \cmidrule(r){3-4} \cmidrule(r){5-6}
& & \boldsymbol{$\alpha=0.1$} & \boldsymbol{$\alpha=0.5$} & \boldsymbol{$\alpha=0.1$} & \boldsymbol{$\alpha=0.5$} \\ \midrule
FedAvg w/o pt & 100 & 67.46 & 79.53 & 44.66 & 51.44  \\ \midrule

FedAvg & 100 & 75.18 & 81.73 & 51.18 & 55.83  \\
FedAvg-RDS & 10 & 75.05 & 81.37 & 50.22 & 53.27   \\
FedProx & 100 & 78.48 & 80.96 & 50.80 & 55.43  \\ 
FedProx-RDS & 10 & 76.41 & 80.28 & 50.18 & 53.44 \\ 
\midrule

\textbf{FedFT-RDS} & 10 & 81.11 & 85.51 & 53.66 & 56.57  \\
\textbf{FedFT-EDS} & 10 & 83.82 \contour{black}{$\uparrow$} & 86.24 \contour{black}{$\uparrow$} & 54.04 \contour{black}{$\uparrow$} & 57.03 \contour{black}{$\uparrow$} \\ \midrule

Centralised & 100 & \multicolumn{2}{c}{87.47} & \multicolumn{2}{c}{58.79}   \\

\bottomrule
\end{tabular}

\label{tab: local data selection CIFAR-10/100, 10 clients.}
\end{table}

Further, Figure~\ref{fig: FedFT ds, CIFAR-10/100 10 clients training curves} depicts the learning curve of FedFT-EDS and that of the baselines during FL iterations. We see that partial model fine-tuning achieves faster convergence than the baseline models even when using 10\% of the available local training data, with FedFT-EDS slightly better than FedFT-RDS. By contrast, FedAvg without pretraining has the worst convergence rate, which makes training from scratch inefficient for devices with reduced computational resources. 

\begin{figure}
  \centering
    \begin{subfigure}{.48\linewidth}
    \centering
    \centerline{\includegraphics[width=\linewidth]{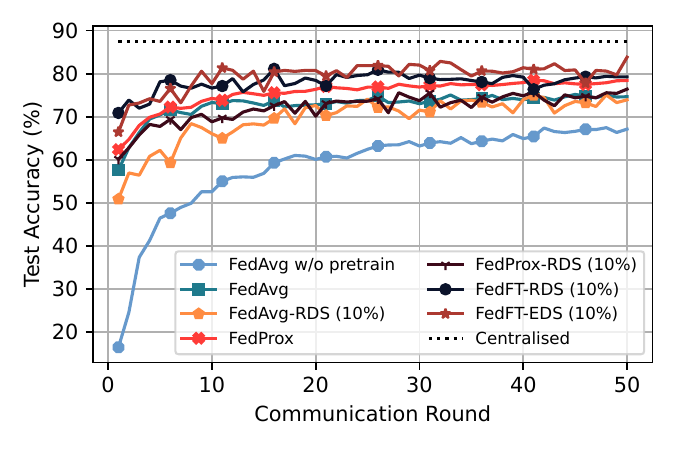}}
    \caption{CIFAR-10, $Diri(0.1)$}
    \end{subfigure} 
    \begin{subfigure}{.48\linewidth}
    \centering
    \centerline{\includegraphics[width=\linewidth]{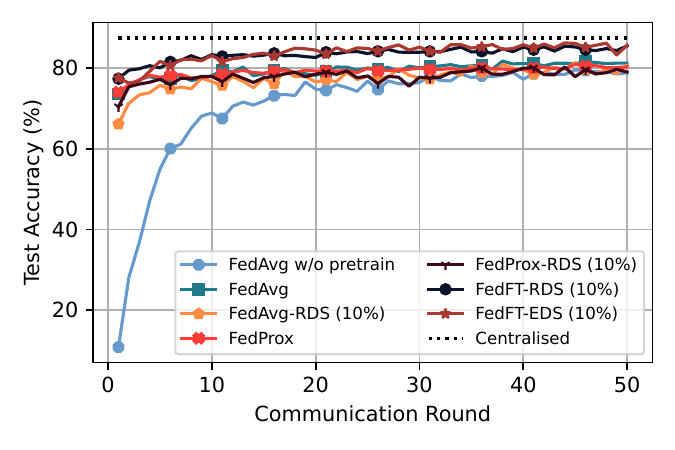}}
    \caption{CIFAR-10, $Diri(0.5)$}
    \end{subfigure} 

    \begin{subfigure}{.48\linewidth}
    \centering
    \centerline{\includegraphics[width=\linewidth]{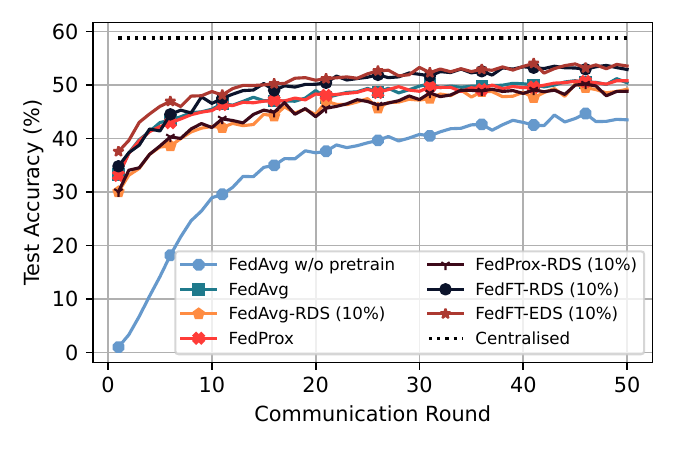}}
    \caption{CIFAR-100, $Diri(0.1)$}
    \end{subfigure} 
    \begin{subfigure}{.48\linewidth}
    \centering
    \centerline{\includegraphics[width=\linewidth]{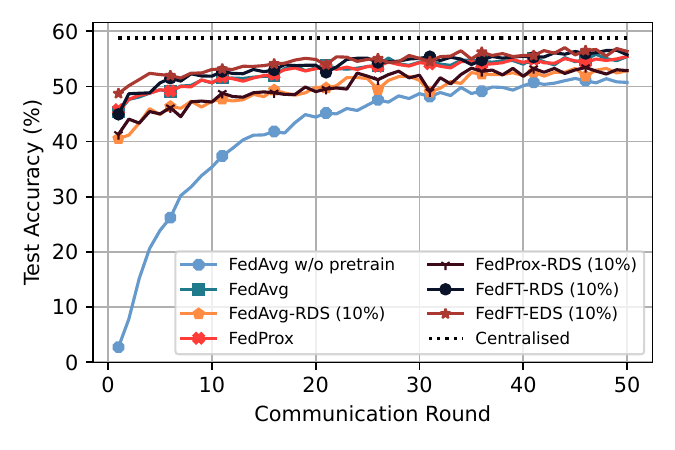}}
    \caption{CIFAR-100, $Diri(0.5)$}
    \end{subfigure} 
    
  \caption{Learning curves of FedFT-EDS and baselines, with global model accuracies computed over test data.}
  \label{fig: FedFT ds, CIFAR-10/100 10 clients training curves}
\end{figure}

\subsection{Improved Learning Efficiency with FedFT-EDS}

Computational cost on small clients is a big issue in FL. We introduce a metric, the learning efficiency, to demonstrate the reduction in computational cost brought by FedFT-EDS.
Specifically, the learning efficiency is calculated by dividing the best test accuracy of the global model by the total training time spent on all participating clients during FL iterations. Through this, we measure the amount of accuracy points gained by the global model from each unit of time spent during local training (seconds). An FL method with high learning efficiency implies that its clients use less training time to achieve the same FL performance, which is more suited for resource-constrained devices. 

Figure~\ref{fig: FedFT, ds, CIFAR-10/100 10 clients training efficiency} compares the learning efficiency of our FedFT-EDS with the chosen baselines. Selecting just 10\% of the client local data for each epoch, FedFT-EDS not only achieves the best absolute global model performance but it also triples the learning efficiency over the baselines (FedAvg, FedProx) on CIFAR-10. Also on CIFAR-100, the learning efficiency is improved by 5$\times$. Although FedAvg-RDS achieves a close learning efficiency to FedFT-EDS, its global performance is severely penalised, up to 10\% on CIFAR-10 and 4\% on CIFAR-100 due to training on less relevant data. 
These results demonstrate the efficacy of entropy-based data selection.

\begin{figure}
  \centering
    \begin{subfigure}{.48\linewidth}
    \centering
    \centerline{\includegraphics[width=\linewidth]{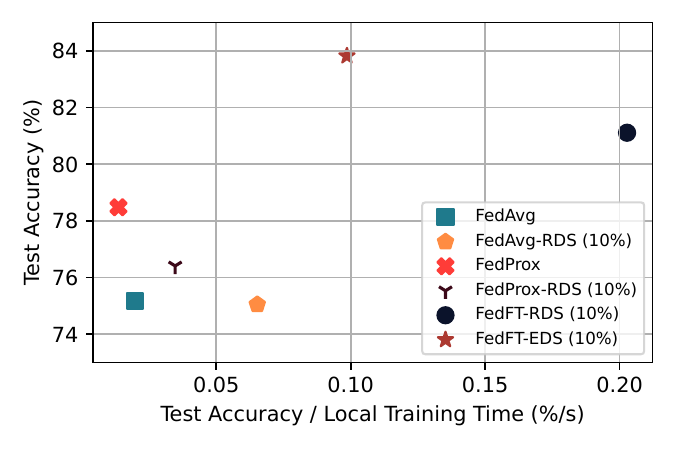}}
    \caption{CIFAR-10, $Diri(0.1)$}
    \end{subfigure} 
    \begin{subfigure}{.48\linewidth}
    \centering
    \centerline{\includegraphics[width=\linewidth]{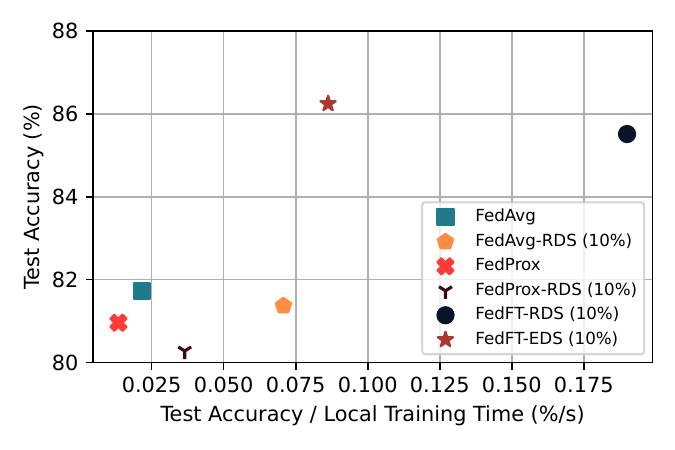}}
    \caption{CIFAR-10, $Diri(0.5)$}
    \end{subfigure} 

    \begin{subfigure}{.48\linewidth}
    \centering
    \centerline{\includegraphics[width=\linewidth]{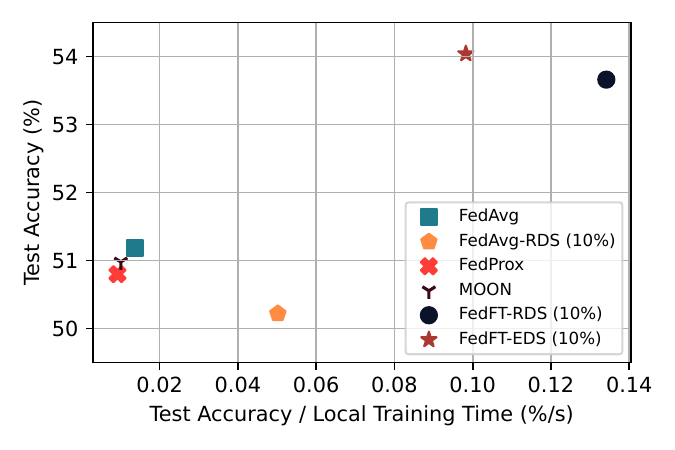}}
    \caption{CIFAR-100, $Diri(0.1)$}
    \end{subfigure} 
    \begin{subfigure}{.48\linewidth}
    \centering
    \centerline{\includegraphics[width=\linewidth]{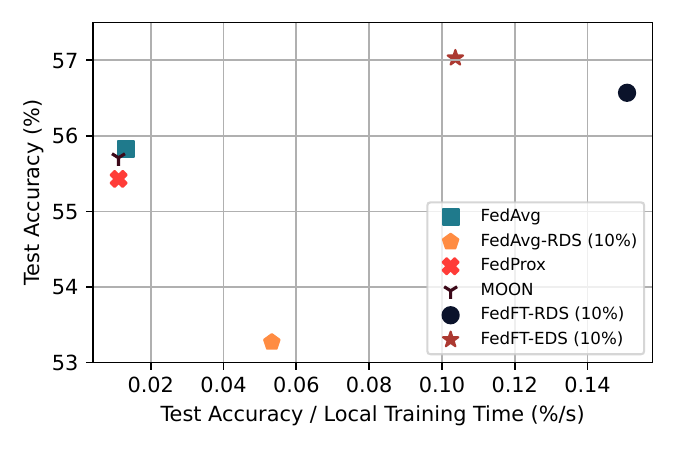}}
    \caption{CIFAR-100, $Diri(0.5)$}
    \end{subfigure} 
    
  \caption{Comparison of learning efficiency, calculated by dividing top test accuracy by total local training time. FedFT-EDS at least triples the learning efficiency of baselines and achieves the best global model performance in all cases, demonstrating significant improvement.}
  \label{fig: FedFT, ds, CIFAR-10/100 10 clients training efficiency}
\end{figure}

\subsection{Overcoming the Straggler Issue with FedFT-EDS}

We further simulate a larger client pool with stragglers dropping out in FedAvg to replicate some clients folding under the standard FL heavy workload. Thus, we show how \mbox{FedFT-EDS} improves FL by addressing the straggler issue through lightweight workloads. 
In this experiment, we set the number of clients to 100. The fraction of participating clients in FedAvg is indicated by $f_{n}$, while the remaining $1-f_{n}$ is the fraction of clients that become stragglers. In contrast, FedFT-EDS assumes full client participation because the adopted partial model fine-tuning and training on a reduced amount of data lower the client effort of updating the model. 
We compare FedFT-EDS with FedAvg given three client participation ratios by choosing $f_{n}$ to be 100\%, 20\%, and 10\%, thus simulating the entire range of conditions from full client participation to low client participation. Moreover, we are also interested in understanding the effect of increasing the volume of training data selected in FedFT-EDS. Therefore, we explore the case of 50\% local data being selected in both entropy-based data selection and random data selection. 

Table~\ref{tab: FedFT, ds, local data selection CIFAR-10/100, 100 clients.} presents the results of running the training over 100 clients as presented above. These results demonstrate that FedFT-EDS outperforms FedAvg with full client participation, even in the case of a larger client pool for the latter. For realistic conditions of straggler dropout for FedAvg, the performance gap is much enlarger, up to 7\% difference under strong data heterogeneity for $Diri(0.1)$. The significant performance boost achieved by FedFT-EDS also exposes the importance of allowing contributions from  all clients by reducing their workload. Similar to our previous experiments with 10 clients, here again we see FedFT-EDS outperforming FedFT-RDS in all scenarios with 100 clients, highlighting the efficacy of entropy-based data selection in improving the global model performance. 

Looking at the amount of training instances (Table~\ref{tab: FedFT, ds, local data selection CIFAR-10/100, 100 clients.}), we see that the difference in global model performance between selecting 10\% of local instances and selecting 50\% of local instances is of about 1\% on CIFAR-10 and 3.5\% on CIFAR-100 in favor of the latter, for both FedFT-RDS and FedFT-EDS. 
However, more training data is not always better, as we see that the baseline FedFT-ALL using all the available local data for local model update actually harms FL performance compared to FedFT-EDS with 50\% active data selection. 
This observation articulates the critical hypothesis made in this work -- not all client data is beneficial for FL and entropy-based data selection is able to filter out divergent client data. This insight is critical to advancing FL and particularly useful for setups with resource-constrained clients. 

\begin{table*}
\caption{Top-1 accuracy (\%) of FL with 100 clients scenario with straggler simulations. All variants of FedFT assume full client participation as their partial model fine-tuning are highly efficient. A critical finding is not all local data is beneficial for federated learning as FedFT-EDS with 50\% outperforms FedFT-ALL.}
\centering
\begin{tabular}{@{}lllllll@{}}
\toprule
\multirow{2}{*}{\textbf{Method}} & \multirow{2}{*}{\boldsymbol{$f_{n}$}} & \multirow{2}{*}{\boldsymbol{$P_{ds}$}} & \multicolumn{2}{c}{\textbf{CIFAR-10}} & \multicolumn{2}{c}{\textbf{CIFAR-100}}  \\ \cmidrule(r){4-5} \cmidrule(r){6-7}
& & & \boldsymbol{$\alpha=0.1$} & \boldsymbol{$\alpha=0.5$} & \boldsymbol{$\alpha=0.1$} & \boldsymbol{$\alpha=0.5$} \\ \midrule
FedAvg w/o pret.  & 100\% & 100\% & 55.79 & 72.00 & 25.97 & 30.66  \\ \midrule

FedAvg & 100\% & 100\% & 77.54 & 80.00 & 46.60 & 49.78  \\
FedAvg  & 20\% & 100\% & 77.03 & 80.77 & 45.94 & 49.80  \\ 
FedAvg  & 10\% & 100\% & 75.20 & 80.49 & 44.17 & 49.20  \\ \midrule

\textbf{FedFT-RDS} & 100\% & 10\% & 78.20 & 80.25 & 47.64 & 50.23   \\
\textbf{FedFT-EDS} & 100\% & 10\% & 78.92 \contour{black}{$\uparrow$}& 81.74 \contour{black}{$\uparrow$} & 48.22 \contour{black}{$\uparrow$}& 50.74 \contour{black}{$\uparrow$}  \\ \midrule

\textbf{FedFT-ALL} & 100\% & 100\% & 78.96 & 81.26 & 50.39 & 53.23   \\
\textbf{FedFT-RDS} & 100\% & 50\% & 79.02 & 81.57 & 50.51 & 53.33   \\
\textbf{FedFT-EDS} & 100\% & 50\% & 79.80 \contour{black}{$\uparrow$}& 82.46 \contour{black}{$\uparrow$} & 51.54 \contour{black}{$\uparrow$}& 54.22 \contour{black}{$\uparrow$}  \\ 

\bottomrule
\end{tabular}

\label{tab: FedFT, ds, local data selection CIFAR-10/100, 100 clients.}
\end{table*}

Figure~\ref{fig: FedFT, ds, CIFAR-10/100 100 clients training efficiency} presents the learning efficiency metric introduce above computed for the scenario of 100 clients. FedFT-EDS consistently exhibits better learning efficiency and achieves best global model performance. FedFT-EDS has 3.5\% better accuracy over FedFT-RDS, but slightly lower learning efficiency due to overhead in local data entropy calculation. 

\begin{figure}
  \centering
    \begin{subfigure}{.48\linewidth}
    \centering
    \centerline{\includegraphics[width=\linewidth]{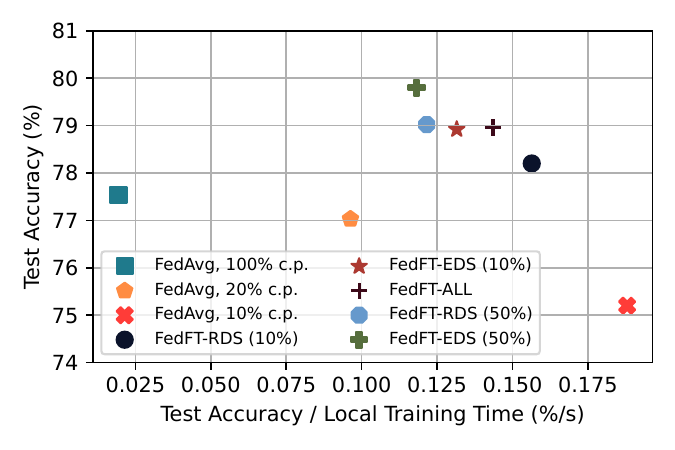}}
    \caption{CIFAR-10, $Diri(0.1)$}
    \end{subfigure} 
    \begin{subfigure}{.48\linewidth}
    \centering
    \centerline{\includegraphics[width=\linewidth]{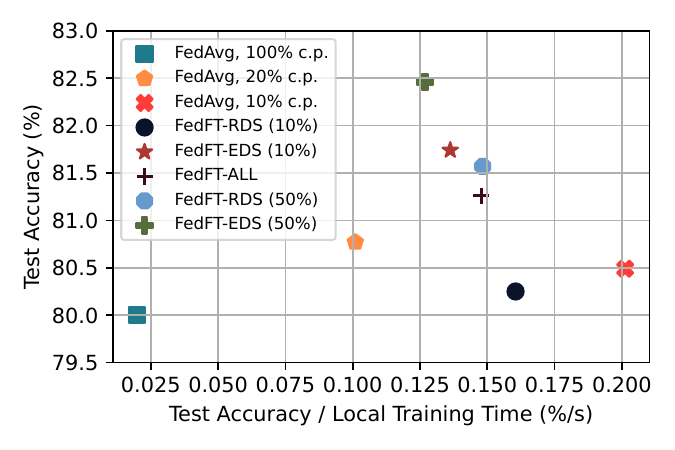}}
    \caption{CIFAR-10, $Diri(0.5)$}
    \end{subfigure} 

    \begin{subfigure}{.48\linewidth}
    \centering
    \centerline{\includegraphics[width=\linewidth]{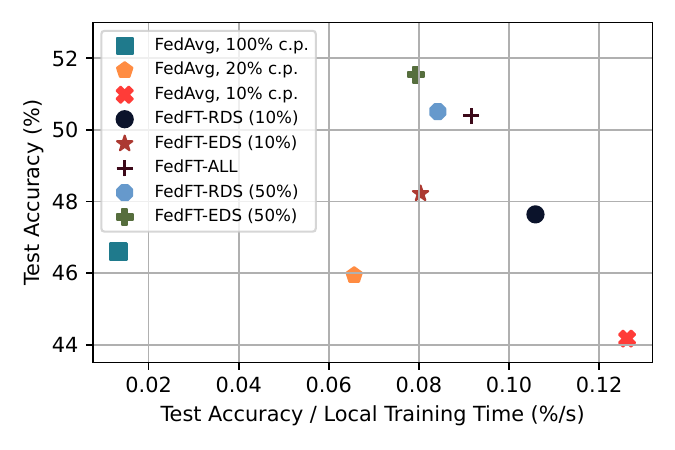}}
    \caption{CIFAR-100, $Diri(0.1)$}
    \end{subfigure} 
    \begin{subfigure}{.48\linewidth}
    \centering
    \centerline{\includegraphics[width=\linewidth]{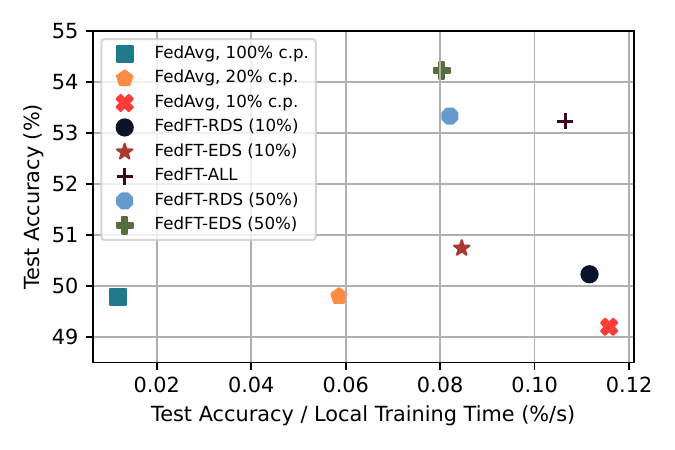}}
    \caption{CIFAR-100, $Diri(0.5)$}
    \end{subfigure} 
    
  \caption{The learning efficiency of FedFT-EDS, FedFT-RDS, FedFT-ALL, and the baselines in the 100-client scenario. FedFT-EDS (50\%) trades a small amount of learning efficiency to achieve the best global model performance. While FedAvg achieves the best learning efficiency with 10\% client participation, its global model performance is significantly compromised by the straggler issue.}
  \label{fig: FedFT, ds, CIFAR-10/100 100 clients training efficiency}
\end{figure}

\subsection{Improved Global Model Convergence with FedFT-EDS}

We also see a faster convergence of FedFT-EDS over FedFT-RDS. Figure~\ref{fig: fedFT, ds, CIFAR-10/100 100 clients training curves} and Figure~\ref{fig: FedFT, ds, CIFAR-10/100 100 clients training curves, compare rds and eds} present the learning curves during the FL training rounds, where FedFT-EDS outperforms all the other methods, especially on the first few rounds. From round 20, the other methods manage to close the gap, but in all evaluated conditions our FedFT-EDS outperforms them in global model performance. These results offer a clear indication of the importance of selecting the right training samples early, which entropy based selection manages to produce.

Effectively, our proposed solution FedFT-EDS accelerates the training in FL, which is another advantage when running on resource-constrained devices since participating in just a few rounds suffices for an effective training.

\begin{figure}
  \centering
    \begin{subfigure}{.48\linewidth}
    \centering
    \centerline{\includegraphics[width=\linewidth]{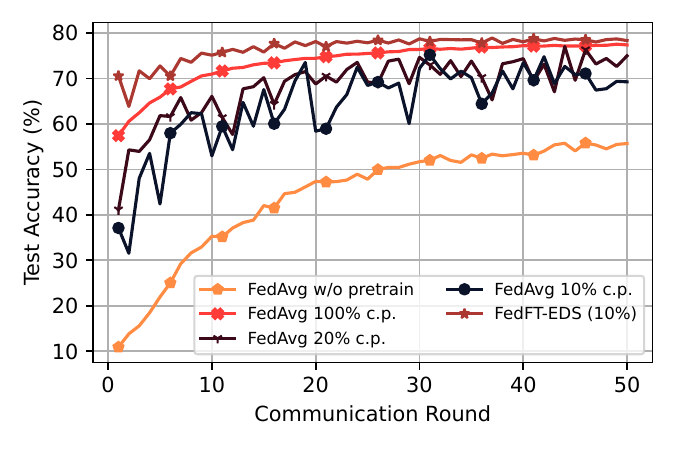}}
    \caption{CIFAR-10, $Diri(0.1)$}
    \end{subfigure} 
    \begin{subfigure}{.48\linewidth}
    \centering
    \centerline{\includegraphics[width=\linewidth]{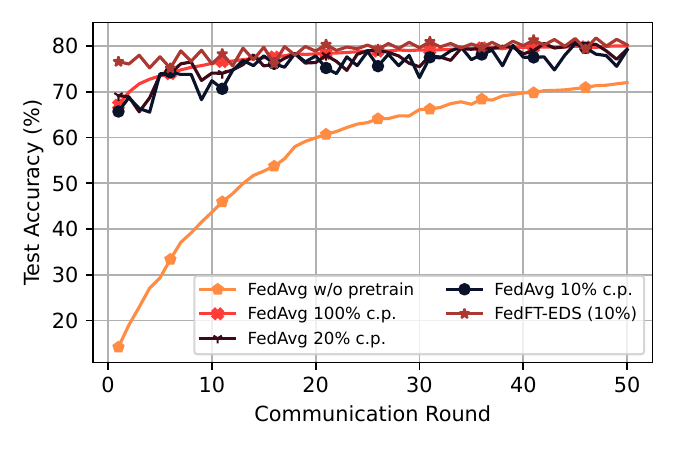}}
    \caption{CIFAR-10, $Diri(0.5)$}
    \end{subfigure} 

    \begin{subfigure}{.48\linewidth}
    \centering
    \centerline{\includegraphics[width=\linewidth]{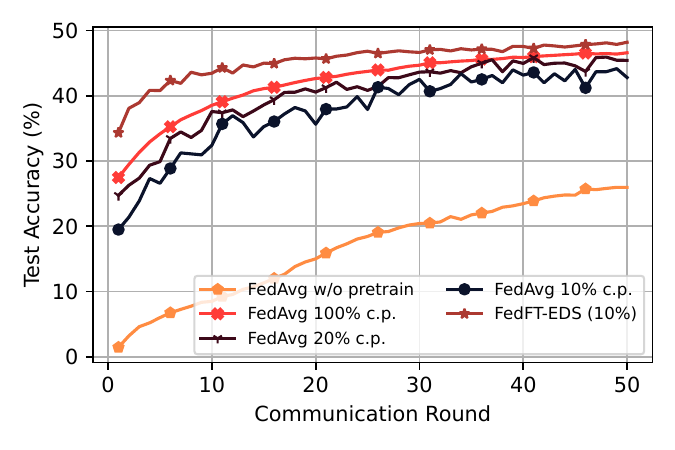}}
    \caption{CIFAR-100, $Diri(0.1)$}
    \end{subfigure} 
    \begin{subfigure}{.48\linewidth}
    \centering
    \centerline{\includegraphics[width=\linewidth]{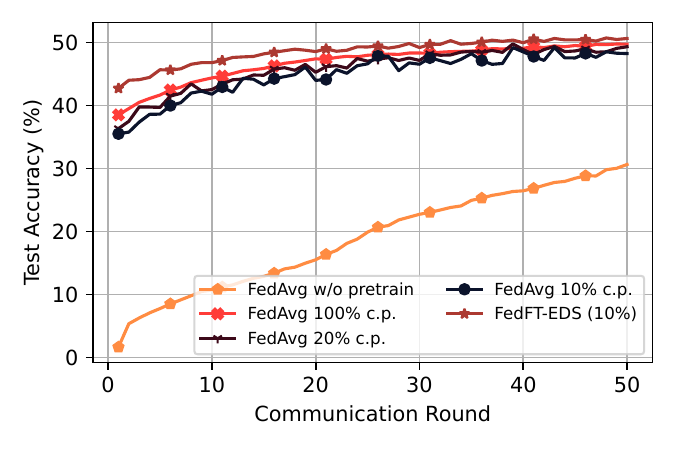}}
    \caption{CIFAR-100, $Diri(0.5)$}
    \end{subfigure} 
    
  \caption{The learning curves of FedFT-EDS, FedFT-RDS and other baselines. FedFT-EDS consistently outperforms all baselines throughout FL rounds.}
  \label{fig: fedFT, ds, CIFAR-10/100 100 clients training curves}
\end{figure}


\begin{figure}
  \centering
    \begin{subfigure}{.48\linewidth}
    \centering
    \centerline{\includegraphics[width=\linewidth]{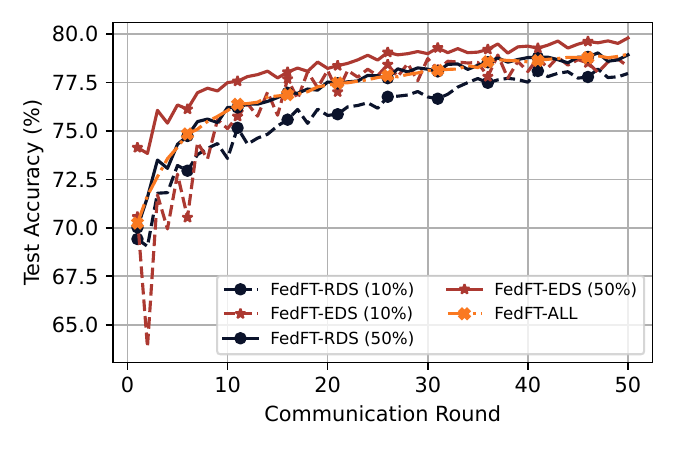}}
    \caption{CIFAR-10, $Diri(0.1)$}
    \end{subfigure} 
    \begin{subfigure}{.48\linewidth}
    \centering
    \centerline{\includegraphics[width=\linewidth]{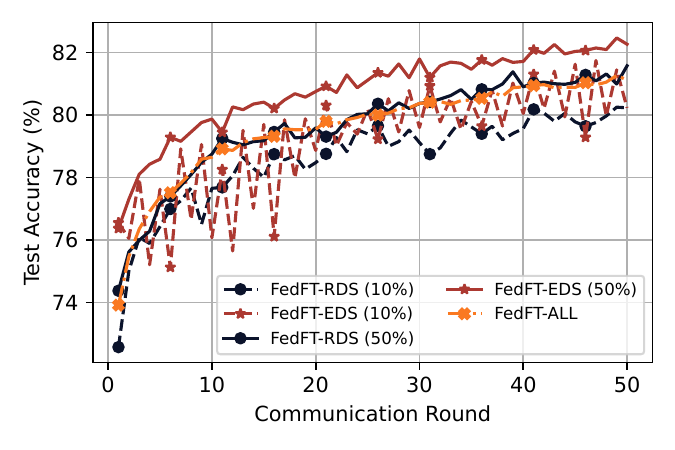}}
    \caption{CIFAR-10, $Diri(0.5)$}
    \end{subfigure} 

    \begin{subfigure}{.48\linewidth}
    \centering
    \centerline{\includegraphics[width=\linewidth]{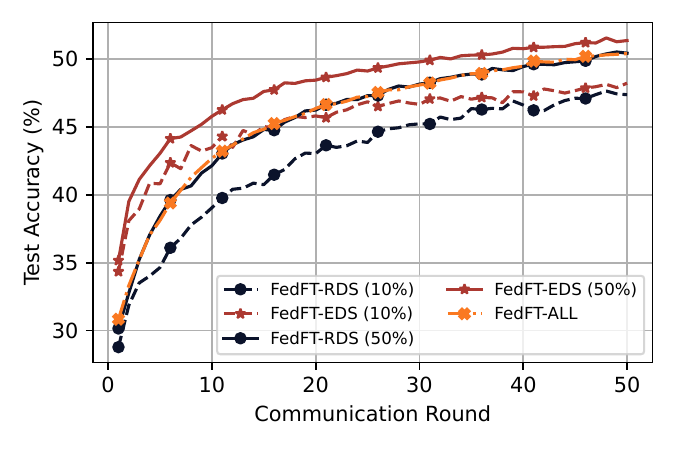}}
    \caption{CIFAR-100, $Diri(0.1)$}
    \end{subfigure} 
    \begin{subfigure}{.48\linewidth}
    \centering
    \centerline{\includegraphics[width=\linewidth]{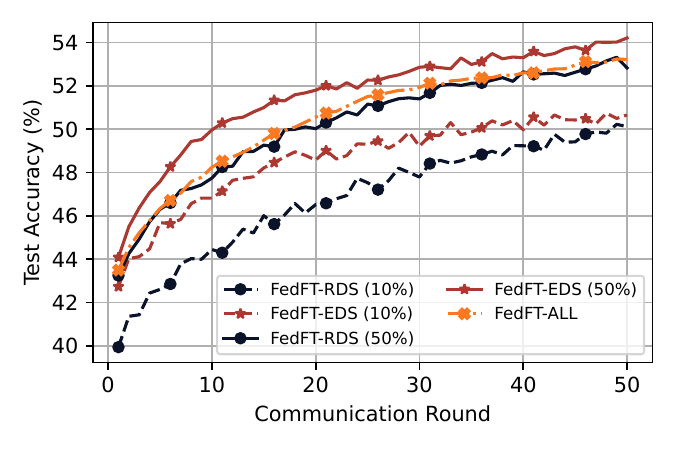}}
    \caption{CIFAR-100, $Diri(0.5)$}
    \end{subfigure} 
    
  \caption{Varying the amount of training data selection for FedFT-EDS and FedFT-RDS, with 10\% and 50\% of local data. In both cases, FedFT-EDS outperforms FedFT-RDS significantly in convergence and top test accuracy.}
  \label{fig: FedFT, ds, CIFAR-10/100 100 clients training curves, compare rds and eds}
\end{figure}

\subsection{Cross-Domain Evaluation}
Besides image classification, we extend the evaluation of FedFT-EDS to another cross-domain scenario, in speech recognition using the Google Speech Command dataset. In this scenario, we concentrate on the generalisation performance of FedFT-EDS, evaluating the efficacy of partial model fine-tuning and entropy-based data selection on the target domain, which is significantly different from the pretraining domain. We assume a full client participation setting with 100 clients as before, and set a strong data heterogeneity with $Diri(0.1)$. 

Table~\ref{tab: FedFT, ds, local data selection GSC, 100 clients.} presents the global model accuracy (\%) achieved by the vanilla FedAvg without pretraining, FedAvg with pretraining the global model, FedFT-RDS, and our FedFT-EDS. Additionally, we report the performance of centralised learning as the upper bound achievable on the dataset. The observation is twofold. First, pretraining the global model is still beneficial across domains, improving the performance of FedAvg considerably by over 14\%. This again shows that pretraining mitigates the model shift on the client side in FL. This holds even across domains, so we generalise this insight so advice on pretraining from other domains when data in the target domain is scarce, instead of starting FL from scratch. Though future studies on which source domain is most relevant for the target domain might give more practical guidance of domain selection. Second, the entropy-based data selection achieves better performance over random data selection even for the cross-domain scenario. Given 50\% data selected for updating the client models, FedFT-EDS outperforms FedFT-RDS by 4\%. This confirms our earlier observation about the advantage of entropy-based data selection, here showing effectiveness even in different domains, between pretraining task and downstream learning task.

\begin{table}
\caption{Top-1 accuracy (\%) of FedFT on GSC, using Small ImageNet for the pretraining phase. Pretraining and fine-tuning are still effective even in the strong domain shift case. The entropy data selection is more effective for improving the model performance than random data selection regardless of the domain shift.}

\centering
\begin{tabular}{@{}lll@{}}
\toprule

\textbf{Method} & \boldsymbol{$P_{ds}$} & \textbf{Top-1 Acc}  \\ \midrule

FedAvg w/o pt. & 100\% &  41.33 \\ 
FedAvg w/ pt.  & 100\% & 55.40  \\ \midrule

\textbf{FedFT-RDS} & 10\% & 54.96   \\
\textbf{FedFT-EDS} & 10\% & 55.60 \contour{black}{$\uparrow$} \\  \midrule
\textbf{FedFT-RDS} & 50\% & 55.56   \\
\textbf{FedFT-EDS} & 50\% & 59.32 \contour{black}{$\uparrow$} \\ \midrule

Centralised learning & 100\% & 91.22 \\
\bottomrule
\end{tabular}


\label{tab: FedFT, ds, local data selection GSC, 100 clients.}
\end{table}

\subsection{Ablation Studies}\label{sec: ablation study}

In this ablation study we investigate further the impacts of some important factors on FedFT-EDS. These are the level of data heterogeneity defined by $Diri(\alpha)$, the depth of the partial model used for fine-tuning, and the temperature for the hardened softmax. We use CIFAR-100 for this ablation study, with 100 clients in full participation, taking advantage of the lower workload of FedFT-EDS. The proportion of selected client data ($P_{ds}$) is set to 50\%. To ground the observations, we also include the results for FedFT-RDS using random data selection. 

\begin{figure}
  \begin{center}
    \includegraphics[width=0.95\linewidth]{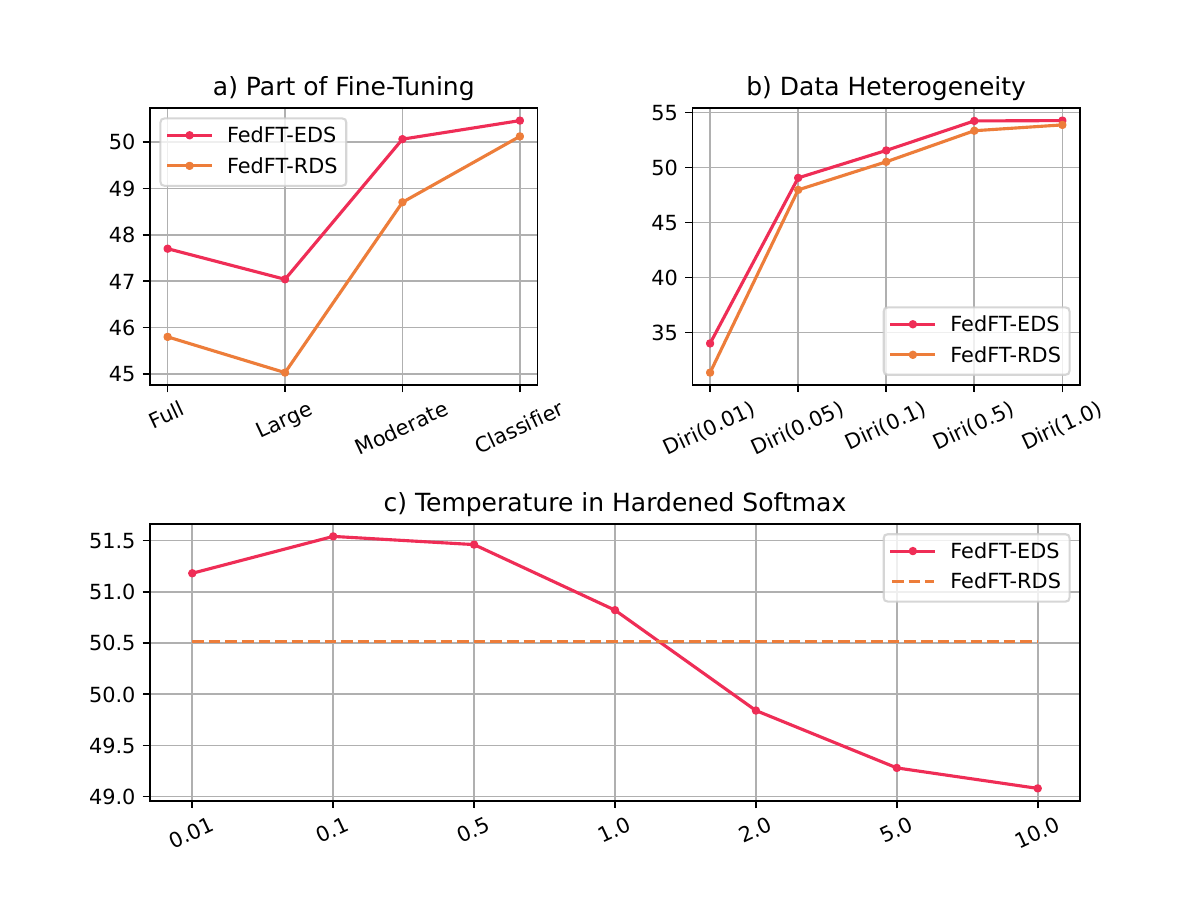}
  \end{center}
  \caption{Ablation study of FedFT-EDS: a) Model part to fine-tuning. b) Data heterogeneity c) Temperature in the hardened softmax.}
  \label{fig: ablation study}
\end{figure}

\subsubsection{Fine-tuning different model parts}
After pretraining, FedFT-EDS allows us to fine-tune just a smaller part at the top of the model during FL. Consequently, the performance of FedFT-EDS is decided by the size of this trainable (fine-tune) partial of the model. Here, we explore how the performance of FedFT-EDS varies with the size of the trainable partial of model. Effectively, we compare fine-tuning the entire model, fine-tuning a larger part of the model, by fixing only the bottom layer group, a medium-size part of model, by fixing both bottom and middle layer groups, and only the classifier by fixing all the layer groups below the classifier layer. We use $Diri(0.1)$ to simulate client data heterogeneity. Figure~\ref{fig: ablation study} a) shows that the performance of FedFT-EDS and FedFT-RDS with these predefined fixes in the model size. Most insightful observation is that training a larger part of the model yields worse performance. 
Conversely, the best performance is achieved by just fine-tuning the classifier part of the model. 
However, we defend this conclusion only when the pretraining domain is very similar to the target domain, so that shareable features appear in the lower part of the model.
The trainable size of the global model should be determined based on the similarity between source domain and target domain. 

In all scenarios, FedFT-EDS outperforms FedFT-RDS. Moreover, we see that the gap increases with more trainable layers, nearly 2\% when train the entire model or a larger part of the model. This observation indicates that entropy information guides the selection of samples to those instances that contribute the most to updating many of the parameters in the right direction. 



\subsubsection{Varying data heterogeneity levels}
We now check if data heterogeneity level plays a role in the performance achieved by entropy-based data selection. Particularly, we set $\alpha$ in the Dirichlet distribution 0.01, 0.05, 0.1, 0.5, and 1.0 to simulate scenarios of both strong and weak data heterogeneity. Figure~\ref{fig: ablation study} b) shows that FedFT-EDS consistently outperforms FedFT-RDS across all scenarios. Larger performance gaps are found at scenarios with strong data heterogeneity, with reduced performance gain using entropy-based data selection when client data is approaching IID. This is a strong evidence that training the distributed global model on instances selected by entropy is helpful to mitigate the model shift which is attributed to the data heterogeneity and can penalise FL performance. 

\subsubsection{Using different temperatures in hardened softmax}
The temperature $\rho$ in the softmax activation determines the effectiveness of the proposed entropy-based data selection. Our FedFT-EDS introduces the hardened softmax to calculate the entropy of client data, choosing $\rho<1$, which is capable of determining the more difficult instances to learn. Figure~\ref{fig: ablation study} presents the performance of FedFT-EDS when using different temperature $\rho$ values, ranging from 0.01 to 10. The baseline performance is the random data selection, FedFT-RDS. As before, $Diri(0.1)$ is used to control client data heterogeneity. 

Values of $\rho$ smaller than 1.0 bring a clear advantage to entropy-based data selection over random data selection. When we soften the softmax activation by setting $\rho > 1.0$, FedFT-EDS is outperformed by FedFT-RDS. Our explanation for this is that using a softened softmax to calculate entropy and rank training data instances causes easier-to-learn samples to blend into the selected data set, as their entropy is increased by the softened softmax. This observation reinforces our claim that not all client data is beneficial for FL. When samples with low learning value are selected for local updates, FL with active data selection may perform worse than random data selection. 

\section{Conclusion}

We presented FedFT-EDS, a novel approach to enhance Federated Learning (FL) by integrating partial model fine-tuning with entropy-based data selection. By determining the most informative data points for training, FedFT-EDS effectively reduces the training workload on resource-constrained devices while improving the overall efficiency and performance of the global model. Experiments on CIFAR-10 and CIFAR-100 demonstrated that FedFT-EDS uses only 50\% of client data, accelerates training by up to three-fold, and outperforms established methods like FedAvg and FedProx in heterogeneous settings. Our findings highlight the importance of targeted data selection in FL systems, paving the way for more scalable, resource-efficient learning frameworks.

\balance

\bibliographystyle{IEEEtranS}
\bibliography{main}

\end{document}